\documentclass[10pt,twocolumn,letterpaper]{article}
\usepackage[accsupp]{axessibility}
\usepackage{cvpr}      
\definecolor{cvprblue}{rgb}{0.21,0.49,0.74}
\usepackage[pagebackref,breaklinks,colorlinks,allcolors=cvprblue]{hyperref}

\def\paperID{V4A-44} 
\def\confName{CVPR}
\def\confYear{2026}

\title{\ FruitProM-V2: Robust Probabilistic Maturity Estimation and Detection of Fruits and Vegetables }

\author{Rahul Harsha Cheppally \and Sidharth Rai \and Sudan Baral \and Benjamin Vail \and Ajay Sharda\\
Kansas State University, Manhattan, Kansas, USA\\
{\tt\small \{r4hul, sidharth, sudanb, benv86, asharda\}@ksu.edu}
}

\begin{document}
\maketitle
\begin{abstract}
Accurate fruit maturity identification is essential for determining harvest timing, as incorrect assessment directly affects yield and post-harvest quality. Although ripening is a continuous biological process, vision-based maturity estimation is typically formulated as a multi-class classification task, which imposes sharp boundaries between visually similar stages. To examine this limitation, we perform an annotation reliability study with two independent annotators on a held-out tomato dataset and observe disagreement concentrated near adjacent maturity stages. Motivated by this observation, we model maturity as a latent continuous variable and predict it probabilistically using a distributional detection head, converting the distribution into class probabilities through the cumulative distribution function (CDF). The proposed formulation maintains comparable performance to a standard detector under clean labels while better representing uncertainty. Furthermore, when controlled label noise is introduced during training, the probabilistic model demonstrates improved robustness relative to the baseline, indicating that explicitly modeling maturity uncertainty leads to more reliable visual maturity estimation.
\end{abstract}    
\section{Introduction}
\label{sec:intro}
Assessment of fruit ripeness is a critical requirement for determining the correct harvest time \cite{Prasanna2007-mu}. Harvesting prematurely reduces flavor and market value, whereas overripe fruit becomes soft and susceptible to tissue damage during handling and transportation \cite{Taiti2025-ki}. Consequently, reliable maturity estimation directly impacts post-harvest quality, storage life, and supply-chain losses. Historically, ripeness has been evaluated manually, where the harvester judges maturity using subjective visual and tactile cues. Workers typically rely on skin color, surface appearance, aroma, and perceived firmness to assign a maturity stage. While this procedure is rapid and requires no instrumentation, it depends heavily on individual experience and environmental conditions, resulting in substantial inter-observer variability. Post-harvest studies indicate that these external cues correlate imperfectly with internal physiological attributes such as soluble solid content, starch degradation, and ethylene activity, which more accurately characterize ripening \cite{Abbott1999QualityMeasurement,Kader2002Postharvest}. As a result, identical fruit samples may be graded differently by different workers, producing inconsistent harvest decisions and variable quality during storage and transport \cite{Peirs2002AppleNIR}. These limitations motivate the need for objective and repeatable sensing methods, particularly computer vision approaches capable of estimating ripeness directly from observable phenotypic characteristics while reducing human bias and improving harvesting consistency.

Beyond manual grading, numerous techniques have been proposed to estimate fruit maturity using objective measurements. Conventional laboratory assessments—such as total soluble solid content (TSS), titratable acidity, dry matter analysis, and firmness testing using penetrometers—provide strong correlation with physiological ripeness but require cutting or puncturing the fruit and therefore cannot be deployed continuously or at scale in field environments \cite{Abbott1999QualityMeasurement,Kader2002Postharvest}. While reliable for post-harvest evaluation, these destructive procedures limit their applicability to robotic harvesting or pre-harvest monitoring.

To avoid sample damage, non-destructive sensing systems have been explored. Spectroscopic approaches, particularly near-infrared (NIR) and visible–NIR spectroscopy, estimate internal attributes by measuring wavelength-dependent absorption associated with molecular bonds in water and carbohydrates. Although effective for predicting firmness and soluble solids, their performance depends on careful calibration and is sensitive to cultivar variation and environmental conditions \cite{Peirs2002AppleNIR}. Hyperspectral imaging extends this idea by capturing both spatial and spectral information, enabling simultaneous estimation of chemical composition and surface properties. However, hyperspectral systems generate high-dimensional data, require controlled illumination, and incur significant computational cost, limiting their real-time use in commercial harvesting settings. Similarly, volatile-based sensing systems such as electronic noses attempt to infer ripeness from emitted gases (e.g., ethylene), but their measurements are strongly affected by temperature, humidity, and background gases, reducing robustness in outdoor environments.

Mechanical proxies of maturity, including acoustic vibration and impact-based firmness sensing, provide indirect indicators of tissue softening but are often cultivar-dependent and may not consistently reflect biochemical ripening processes \cite{Abbott1999QualityMeasurement}. More broadly, recent surveys report that many sensor-based maturity estimation systems require specialized equipment, close-range measurement, or controlled conditions, making them difficult to deploy reliably under field variability .

Machine learning methods have been applied to these measurements using engineered color, texture, or spectral features. Classical models such as support vector machines and random forests can classify maturity under controlled datasets but depend on handcrafted features and often degrade under changing illumination and background conditions. Deep learning models, particularly convolutional neural networks, improve performance by learning representations directly from images, yet they require large annotated datasets and still struggle with variability and data quality in agricultural environments .

These existing approaches reveal a fundamental trade-off: accuracy vs. scale. Laboratory measurements are accurate but destructive, sensor systems are non-destructive but impractical in-field, and feature-based learning is brittle to environmental variation. Because the ripening process manifests visually through color transition, texture change, and surface appearance, camera-based perception offers a scalable alternative capable of non-contact sensing across large growing areas. This motivates computer-vision maturity estimation methods that operate directly on observable phenotypic cues rather than specialized instrumentation.

Recent advances in computer vision have led to widespread adoption of deep learning for fruit maturity estimation. Convolutional neural networks (CNNs) and modern object detection architectures such as YOLOv5, YOLOv7, and YOLOv8 have been successfully applied to classify ripeness stages directly from RGB imagery, typically formulating the problem as a multi-class classification task (e.g., unripe, intermediate, ripe) \cite{Zhang2022TomatoYOLO, Bu2023CitrusYOLO, Xiao2023YOLOv8Fruit}. These models leverage visual cues including color transitions, texture variation, and surface appearance, and have reported high classification accuracy under controlled or laboratory datasets \cite{Mao2023StrawberryCNN, Ahmed2026Review}. 

However, most existing approaches implicitly assume that maturity stages are discrete categories with reliable ground-truth annotations. In practice, fruit ripening is a gradual biological process characterized by continuous physiological and biochemical change rather than sharp class boundaries \cite{Kader2002Postharvest, Zhang2020-fk}. Consequently, adjacent maturity stages often exhibit subtle visual differences, and even expert annotators may disagree when assigning stage labels, particularly near transition boundaries \cite{Beck2020LabelNoiseAgri, Northcutt2021ConfidentLearning}. 

\begin{figure*}[t]
  \centering
  \includegraphics[width=1.0\textwidth]{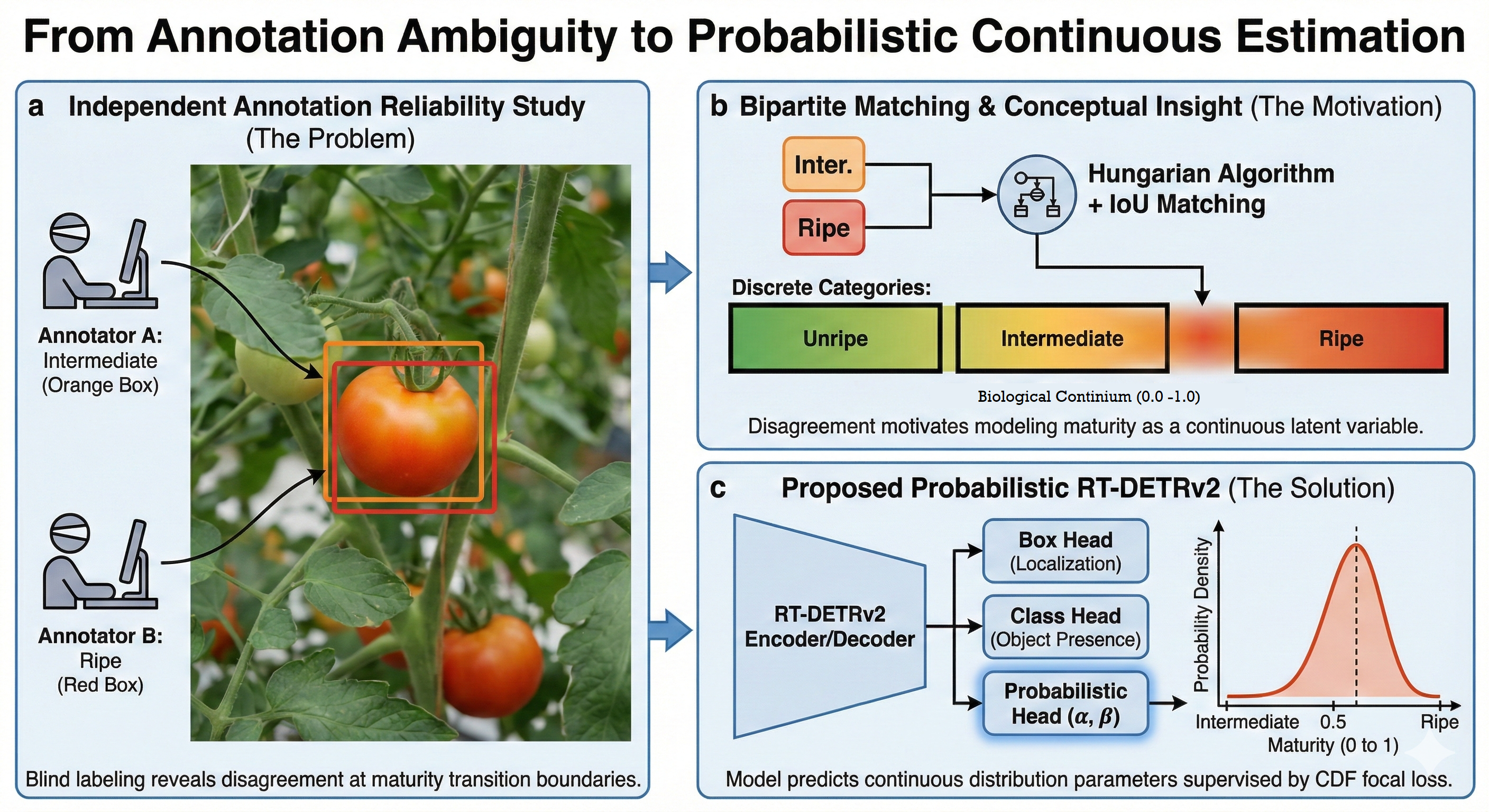}

  \caption{\textbf{From Annotation Ambiguity to Probabilistic Continuous Estimation.} 
  \textbf{(a) Reliability Study:} Our inter-annotator study reveals systematic labeling disagreement on fruits positioned at maturity transition boundaries (e.g., intermediate vs. ripe), suggesting that maturity labels are subjective snapshots of an underlying biological continuum. 
  \textbf{(b) Motivation:} By performing bipartite matching via the Hungarian algorithm, we isolate labeling variance from localization differences. This disagreement justifies modeling maturity as a latent continuous variable rather than a discrete categorical state. 
  \textbf{(c) Proposed Approach:} We extend the RT-DETRv2 architecture with a maturity head that predicts distribution parameters ($\alpha, \beta$). Supervised by a CDF-based focal loss, the model explicitly captures maturity uncertainty and enables high-fidelity continuous ripeness estimation.}
  \label{fig:overview_figure}
\end{figure*}

This discrepancy introduces significant label ambiguity: samples located near stage transitions may receive different categorical labels despite representing similar physiological states. Recent studies further report degradation in model generalization under real orchard conditions due to illumination variability, occlusion, and dataset bias \cite{Kamilaris2018-uq, Zhang2020-fk}. Standard cross-entropy-based classification losses treat labels as certain and mutually exclusive, forcing models to learn sharp decision boundaries for what is inherently a continuous process. As a result, models may achieve high categorical accuracy while failing to capture maturity progression dynamics. These observations suggest that fruit maturity estimation may be more appropriately formulated as an ordinal, probabilistic, or continuous regression problem rather than a deterministic categorical classification task \cite{Geng2016LabelDistributionLearning, Diaz2019OrdinalClassification, Liu2023OrdinalDL}.
In this study, we reformulate fruit maturity estimation as a probabilistic perception task (see Fig.~\ref{fig:overview_figure}). Our contributions are summarized as follows:
\begin{itemize}
    \item We conduct an \textbf{inter-annotator reliability study} that quantifies systematic label ambiguity at maturity transitions, providing empirical evidence that maturity is a latent continuous process.
    \item We propose \textbf{FruitProM} framework that replaces the standard categorical head for maturity detection with a distribution-based head predicting continuous $\alpha$-$\beta$ parameters.
    \item We introduce a \textbf{CDF-based focal loss} that enables robust supervision using interval-based observations, effectively bridging the gap between discrete labels and continuous biological states.
\end{itemize}
Experimental results demonstrate that our formulation better captures maturity progression and maintains high detection performance even under controlled label noise, significantly improving the reliability of vision-based agricultural sensing.

 \section{Related Work}

Modern maturity assessment has transitioned from early CNN foundations—such as feature-concatenation models achieving 0.90 F1 in papaya \cite{Garillos2021}—to high-speed YOLO and Transformer architectures. Real-time frameworks like YOLO-Tomato \cite {Liu2020YOLOTomato} (96.4\% AP) and the parameter-efficient YOLO-DGS \cite{Zhao2025YOLODGS} (80.1\% mAP50) have optimized detection through circular bounding boxes and D-2Detect modules. Despite high precision in strawberries using LW-Swin Transformer backbones \cite{Yang2023} or RT-DETR \cite{Wu2025RTDETR}, these models typically rely on deterministic regression or discrete classification. This formulation ignores the continuous biological ripening described by phenological scales like BBCH \cite{Meier2009BBCH, Bridgemohan2016BBCH}, which resist clean mapping onto discrete categories and lead to structural inter-annotator disagreement \cite{SadeghiTehran2017Wheat}.

To address these limitations, research has integrated uncertainty quantification (UQ) to separate model ignorance (epistemic) from irreducible data noise (aleatoric) \cite{Kendall2017Uncertainties}. While deep ensembles \cite{Lakshminarayanan2017} offer superior calibration, single-pass methods like Evidential Deep Learning \cite{Sensoy2018Evidential, Amini2020DeepEvidential} provide the computational efficiency required for agricultural edge deployment by placing Normal-Inverse-Gamma priors over Gaussian likelihoods. Specialized probabilistic heads, including Mixture Density Networks \cite{Bishop1994} and Beta-distribution regression \cite{Sadowski2019BetaDirichlet}, provide a mathematically grounded alternative to rigid one-hot labels. By leveraging soft probability targets \cite{Hinton2015Distilling, Peterson2019HumanUncertainty} and ordinal-aware noise correction \cite{Garg2023OrdinalNoise}, models can finally account for the biological continuum and irreducible annotation noise, leading to more robust decision-making for harvest timing.

\section{Method}
\label{Method}
\subsection{Problem Formulation}
\label{sec:problem_formulation}
\subsubsection{Standard Detection Assumption}

Given an image $I \in \mathbb{R}^{H \times W \times 3}$, an object
detector predicts a set of $N$ detections
\[
\mathcal{D} = \{(b_i, y_i)\}_{i=1}^{N},
\]
where $b_i \in \mathbb{R}^{4}$ denotes the bounding box of the $i$-th
fruit and $y_i$ is its maturity label. In existing fruit maturity
detection approaches, the maturity state is treated as a categorical
variable:
\[
y_i \in \{1,2,3\},
\]
corresponding to stages such as unripe, intermediate, and ripe.

The detector predicts class probabilities
\[
p(y_i=k \mid I), \quad k\in\{1,2,3\},
\]
and is trained using cross-entropy classification loss.

This formulation assumes that maturity categories are discrete and that
annotations represent the true physiological state. As discussed in
Sec.~\ref{sec:intro}, we instead treat the observed label as a coarse
observation of an underlying latent maturity variable and model maturity
as a continuous probability distribution rather than a deterministic
categorical state.

\subsubsection{Latent Continuous Maturity}

We model fruit maturity as an unobserved continuous variable
\[
m_i \in [0,1],
\]
representing the physiological progression of ripening.

The discrete annotation $y_i$ is a quantized observation of this latent
variable:
\[
y_i =
\begin{cases}
1, & m_i \in [0,\tau_1) \\
2, & m_i \in [\tau_1,\tau_2) \\
3, & m_i \in [\tau_2,1],
\end{cases}
\]
where $\tau_1$ and $\tau_2$ are maturity transition thresholds.
Under this formulation, the label is not the maturity itself but an
interval measurement of the latent biological state.

\subsubsection{Probabilistic Maturity Prediction}

Instead of predicting a discrete class, the detector predicts a
distribution over maturity:
\[
m_i \sim P_\theta(m \mid I, b_i).
\]

We parameterize $P_\theta$ using a Beta distribution:
\[
P_\theta(m \mid I, b_i) = \mathrm{Beta}(m;\alpha_i,\beta_i),
\]
where $(\alpha_i,\beta_i)>0$ are predicted by the network.

The Beta distribution is defined on $[0,1]$ and can represent early,
late, or uncertain maturity states through its shape parameters.

\subsubsection{From Distribution to Class Probabilities}

Human annotations are discrete, so we convert the predicted distribution
into class probabilities using the cumulative distribution function
(CDF):

\[
F(m;\alpha_i,\beta_i) = \int_{0}^{m} \mathrm{Beta}(t;\alpha_i,\beta_i)\,dt.
\]

The probability of each maturity class is obtained by integrating the
distribution over the corresponding interval:

\[
P(y_i=k) =
F(\tau_k;\alpha_i,\beta_i) -
F(\tau_{k-1};\alpha_i,\beta_i),
\]

with $\tau_0=0$ and $\tau_3=1$.

This produces a soft probability vector
\[
\mathbf{p}_i = (p_{i1},p_{i2},p_{i3}), \quad \sum_{k=1}^{3}p_{ik}=1.
\]

Importantly, fruits near maturity boundaries naturally receive
uncertain (distributed) probabilities rather than forced categorical
decisions.

\subsection{Annotation Reliability Study} \label{sec:annotation_study}
To evaluate the reliability of maturity annotations, we conducted an inter-annotator agreement study on the tomato maturity test set. Two independent annotators labeled each visible fruit using the same three maturity categories employed in the original dataset: \textit{Unripe}, \textit{Intermediate}, and \textit{Ripe}. Annotators worked independently and were blinded to both the original dataset annotations and each other's decisions.
\subsubsection{Matching Procedure}
Because object detection produces sets of bounding boxes rather than one-to-one correspondences, direct label comparison requires associating detections across annotation sets. We matched annotated fruits using a bipartite assignment procedure based on bounding box overlap. Given two annotation sets $A = \{a_i\}$ and $B = \{b_j\}$, we computed pairwise intersection-over-union (IoU) scores and solved a minimum-cost matching problem via the Hungarian algorithm~\cite{Kuhn1955-pp, Frank2005-fw}. Only matched pairs with $\text{IoU} \geq \tau = 0.5$ were retained; unmatched detections were excluded from the reliability evaluation.

\subsubsection{Agreement Analysis}
Using the matched pairs, we constructed normalized confusion matrices for three comparisons: (i)~Annotator~A vs.\ Annotator~B, (ii)~Annotator~A vs.\ original dataset labels, and (iii)~Annotator~B vs.\ original dataset labels. This design isolates disagreement in maturity labeling while factoring out localization differences between annotators.

The goal of this study is not to evaluate annotator performance but to characterize the annotation process itself. Systematic disagreement concentrated between adjacent classes would indicate that maturity boundaries are gradual rather than discrete, motivating the probabilistic formulation introduced in Section~\ref{sec:problem_formulation}.

\subsection{Overall Architecture}
\label{sec:architecture}

The proposed framework extends RT-DETR v2~\cite{lv2024rtdetrv2improvedbaselinebagoffreebies}, an
end-to-end real-time object detector, by adding a probabilistic maturity head that
outputs Beta distribution parameters.

\subsubsection{Probabilistic Maturity Head}
\label{sec:maturity_head}

The only architectural change to the base detector is the replacement
of the multi-class classification head with a Probabilistic Maturity
Head. In standard RT-DETR v2, the classification head is an MLP that
maps each query's $d$-dimensional hidden state to $K$ class logits.
We replace this with an MLP of identical structure (three linear
layers, hidden dimension $d = 256$) that instead outputs two scalar
values $(\hat{y}_1, \hat{y}_2)$. These are transformed into the
shape parameters of a Beta distribution:
\[
\alpha = \mathrm{softplus}(\hat{y}_1) + \epsilon, \quad
\beta  = \mathrm{softplus}(\hat{y}_2) + \epsilon,
\]
where $\epsilon = 0.01$ ensures numerical stability. The softplus
activation guarantees $\alpha, \beta > 0$ while providing smooth
gradients throughout the parameter space.

During training, the predicted $(\alpha, \beta)$ are converted to
discrete class probabilities via the Beta CDF, as described in
Sec.~\ref{sec:problem_formulation}, and supervised with focal
loss. During inference, the same CDF integration assigns each
detection to one of three maturity classes. This design means the
network never predicts maturity categories directly; all discrete
class decisions emerge from the underlying continuous distribution.

Bounding box regression and objectness prediction remain unchanged
from the base RT-DETR v2 architecture. Each decoder layer thus
produces three outputs per query: a refined bounding box, an
objectness score, and a maturity distribution
$\mathrm{Beta}(\alpha, \beta)$. This head was supervised at each decoder similar to all the other heads.
\subsubsection{Training with Focal Loss}

We optimize the predicted class probabilities using focal loss \cite{focalloss} rather
than likelihood maximization. For ground-truth label $y_i^{gt}$:
\[
\mathcal{L}_{focal} =
- \sum_{i=1}^{N}
(1 - p_{i,y_i^{gt}})^\gamma
\log(p_{i,y_i^{gt}}),
\]
where $\gamma$ is the focusing parameter.

Focal loss reduces the influence of ambiguous boundary samples and
prevents over-penalizing predictions that are uncertain but reasonable.
This is particularly important because label noise occurs primarily near
maturity transitions.

\subsubsection{Training Objective}
The proposed method replaces only the classification head of the
detector. Bounding box regression and objectness prediction remain
unchanged. The final loss is:
\[
\mathcal{L} =
\mathcal{L}_{box}
+
\mathcal{L}_{obj}
+
\lambda \mathcal{L}_{focal}.
\]
The detector therefore jointly predicts object location and a continuous
representation of fruit maturity.
\subsubsection{Generalizability to Other Architectures}\label{sec:generalizability}
The proposed modification is deliberately minimal: only the classification head is replaced, while the backbone, encoder, decoder, objectness, and bounding box regression remain untouched. This means the approach is not specific to RT-DETR v2. Any object detector whose classification head produces per-detection feature vectors can be adapted in the same way---the feature vector is simply routed through the Beta MLP instead of a softmax classifier, and the focal loss is computed over CDF-derived class probabilities rather than direct logits. This applies equally to anchor-based detectors such as Faster R-CNN~\cite{ren2015faster} and RetinaNet~\cite{focalloss}, anchor-free detectors such as FCOS~\cite{tian2019fcos}, and other DETR variants~\cite{zhu2020deformable, zhang2022dino, huang2026realtimeobjectdetectionmeets, sapkota2509yolo26, sapkota2025rfdetrobjectdetectionvs}. We chose RT-DETR v2 for its strong real-time performance, but the probabilistic maturity formulation is architecture-agnostic and can serve as a drop-in replacement for the classification head in any detection pipeline where the target attribute is better modeled as a continuous quantity observed through discrete labels.
\subsection{Controlled Label Noise Simulation}
To empirically validate the robustness of our proposed probabilistic formulation against the inherent subjectivity of human annotators, we designed a controlled label noise experiment. The inter-annotator reliability study demonstrated that labeling variance is predominantly localized at the transitional boundaries between maturity stages. To simulate this real-world ambiguity under strictly controlled conditions, we introduced artificial symmetric label noise into the training dataset.\par Specifically, we applied a uniform 10\% noise rate to the categorical ground-truth labels. For a randomly selected 10\% subset of the training instances, the original maturity class was uniformly transitioned to an adjacent class, while keeping the bounding box coordinates perfectly intact. This symmetric corruption strategy ensures that the overall spatial distribution of the dataset remains undisturbed, strictly isolating the model's vulnerability to categorical misclassification. The validation and test sets were intentionally kept completely pristine (0\% noise) to ensure all models were evaluated against the same true baseline.
By training the deterministic baselines (YOLOv8l, YOLO11l, and standard RT-DETRv2) and our Probabilistic RT-DETRv2 on this corrupted dataset, we establish a framework to measure how severely standard cross-entropy losses overfit to subjective label errors compared to our continuous distribution-based approach.
\section{Experiments and Results}
\subsection{Implementation Details}
All experiments were conducted on a machine equipped with two NVIDIA RTX 4090 GPUs using CUDA 13.0 and PyTorch. We used the Tomato Maturity dataset from~\cite{Sid2025RTDETRv2}. Since the original dataset provides only training and validation splits, we further divided the validation set into two equal parts (50\%/50\%), using one portion as a validation set and the other as a test set. The validation set was evaluated once per epoch, while the test set was evaluated only once after training. Test results are reported.
\label{sec:results_reliability}

Table~\ref{tab:annotator_agreement} presents the normalized confusion matrices comparing annotators against the baseline and against each other. Both annotators achieved high agreement with the baseline for the polar maturity stages: consistency exceeded 96\% for \textit{Unripe} and 97\% for \textit{Ripe} across both comparisons. However, agreement degraded substantially at the \textit{Intermediate} stage, where both annotators matched the baseline only approximately 48--50\% of the time, with a strong tendency to classify these transition samples as \textit{Ripe} ($\sim$40\%).

Direct inter-annotator comparison confirmed this pattern: while consensus remained robust for \textit{Unripe} (95.8\%) and \textit{Ripe} (84.8\%), agreement on \textit{Intermediate} dropped to 61.4\%. Off-diagonal confusion was predominantly localized to adjacent stages. However, we note an asymmetry in the inter-annotator matrix: while \textit{Unripe} samples were almost never labeled \textit{Ripe} (0.2\%), the reverse confusion---\textit{Ripe} samples labeled \textit{Unripe}---reached 6.4\%. We attribute this to imperfect bounding box association in densely clustered fruit regions, where the Hungarian matching at $\tau = 0.5$ may pair spatially proximate but physiologically distinct fruits. Despite this artifact, the dominant pattern remains clear: disagreement is concentrated at adjacent transition boundaries rather than distributed uniformly across the label space.

\begin{table}[htbp]
\centering
\caption{Normalized confusion matrices for inter-annotator and baseline agreement. Rows indicate the reference label source; columns indicate the comparison target. All values in \%. Matching threshold: IoU $\geq 0.5$.}
\label{tab:annotator_agreement}
\small
\setlength{\tabcolsep}{4pt}
\begin{tabular}{@{}llccc@{}}
\toprule
\textbf{Reference} & \textbf{Class} & \textbf{Unripe} & \textbf{Intermediate} & \textbf{Ripe} \\ \midrule
& & \multicolumn{3}{c}{\textit{vs.\ Annotator 1 (\%)}} \\ \cmidrule(l){3-5}
\textbf{Baseline}
& Unripe        & \textbf{96.9} & 2.6  & 0.5  \\
& Intermediate  & 9.7           & \textbf{47.8} & 42.5 \\
& Ripe          & 0.1           & 0.8  & \textbf{99.1} \\ \midrule
& & \multicolumn{3}{c}{\textit{vs.\ Annotator 2 (\%)}} \\ \cmidrule(l){3-5}
\textbf{Baseline}
& Unripe        & \textbf{98.1} & 1.8  & 0.1  \\
& Intermediate  & 11.5          & \textbf{49.6} & 39.0 \\
& Ripe          & 1.6           & 1.4  & \textbf{97.0} \\ \midrule
& & \multicolumn{3}{c}{\textit{Annotator 1 vs.\ Annotator 2 (\%)}} \\ \cmidrule(l){3-5}
\textbf{Annotator 1}
& Unripe        & \textbf{95.8} & 4.0  & 0.2  \\
& Intermediate  & 21.9          & \textbf{61.4} & 16.8 \\
& Ripe          & 6.4           & 8.8  & \textbf{84.8} \\ \bottomrule
\end{tabular}
\end{table}




2. Model Results: Standard Detection Performance
To establish a performance baseline on the zero-noise dataset, we compared the proposed Probabilistic RT-DETRv2 architecture against the standard deterministic RT-DETRv2 model. Table \ref{tab:map_comparison} summarizes the primary object detection and classification metrics on the validation set.

\begin{table}[ht]
  \caption{Comparison of Mean Average Precision (mAP) across models on the clean Tomato Maturity dataset. Best results are \textbf{bolded}.}
  \label{tab:map_comparison}
  \centering
  \begin{tabular}{@{}lcc@{}}
    \toprule
    Model & mAP\textsubscript{50} & mAP\textsubscript{50--95} \\
    \midrule
    YOLOv8l & 0.962 & 0.866 \\
    YOLOv10x & 0.965 & \textbf{0.874} \\
    YOLO11l & 0.967 & 0.873 \\
    YOLO11x & \textbf{0.968} & 0.866 \\
    YOLO26l & 0.965 & 0.871 \\
    \midrule
    RT-DETRv2 (Baseline) & 0.960 & 0.847 \\
    FruitProm (Ours) & 0.955 & 0.845 \\
    \bottomrule
  \end{tabular}
\end{table}

As shown in Table \ref{tab:map_comparison}, substituting the standard categorical classification head with our continuous probabilistic formulation maintains the fundamental detection capabilities of the architecture. On the unmodified (zero-noise) dataset, the standard baseline achieves a slightly higher overall Average Precision (AP at IoU 0.50:0.95) of 0.847 compared to our probabilistic model's 0.845. The deterministic baseline also shows a marginal advantage in AP@0.50 (0.960 vs 0.955).

However, the proposed probabilistic model demonstrates highly competitive localization precision, slightly outperforming the baseline at the stricter AP@0.75 threshold (0.926 vs 0.923). Furthermore, our model achieves a higher overall Average Recall (0.903 vs 0.899 at 100 maximum detections), indicating a robust ability to successfully localize instances across the maturity spectrum. These results confirm that introducing a probabilistic distribution head does not significantly degrade baseline detection fidelity on pristine data, establishing a stable foundation to evaluate performance under conditions of label ambiguity.
\subsection{Impact of Controlled Label Noise}
Table \ref{tab:final_noise_drop} presents the comparative object detection performance of all evaluated architectures under the 10\% symmetric label noise condition.

\begin{table}[ht]
  \caption{Comparison of robustness under 10\% noise on the Tomato Maturity dataset. Best results are \textbf{bolded}.}
  \label{tab:final_noise_drop}
  \centering
  \small
  \setlength{\tabcolsep}{3pt} 
  \begin{tabular}{@{}lcccc@{}}
    \toprule
    Model & Clean mAP & Noise mAP & Abs. Drop & \% Drop \\
    \midrule
    YOLOv10x & \textbf{0.874} & 0.836 & -0.038 & -4.35\% \\
    YOLO11l & 0.873 & 0.834 & -0.039 & -4.47\% \\
    YOLO26l & 0.871 & 0.833 & -0.038 & -4.36\% \\
    YOLO11x & 0.866 & 0.834 & -0.032 & -3.70\% \\
    YOLOv8l & 0.866 & 0.835 & -0.031 & -3.58\% \\
    \midrule
    RT-DETR\_v2 & 0.847 & 0.821 & -0.026 & -3.07\% \\
    Ours & 0.845 & \textbf{0.840} & \textbf{-0.005} & \textbf{-0.59\%} \\
    \bottomrule
  \end{tabular}
\end{table}

On the pristine zero-noise dataset, the deterministic YOLO11l and YOLOv8l models achieved the highest overall mAP (50-95) at 0.873 and 0.866, respectively. However, the introduction of label noise resulted in substantial performance degradation for all standard deterministic models. YOLO11l experienced the most severe decline, recording an absolute mAP drop of 0.039, which translates to a 4.47\% performance loss. The YOLOv8l and standard RT-DETR baselines exhibited similar vulnerabilities, recording absolute drops of 0.031 (-3.58\%) and 0.026 (-3.07\%), respectively.

Conversely, the proposed Fruitprom architecture maintained highly stable performance under the noisy condition. The model recorded a marginal absolute drop of 0.005 (-0.59\%), yielding a final mAP of 0.840. As a result, Fruitprom achieved the highest absolute performance in the 10\% noise regime, surpassing the deterministic RT-DETR baseline (0.821) as well as the YOLOv8l (0.835) and YOLO11l (0.834) architectures.

\begin{figure*}[t] 
  \centering
  \includegraphics[width=0.32\linewidth]{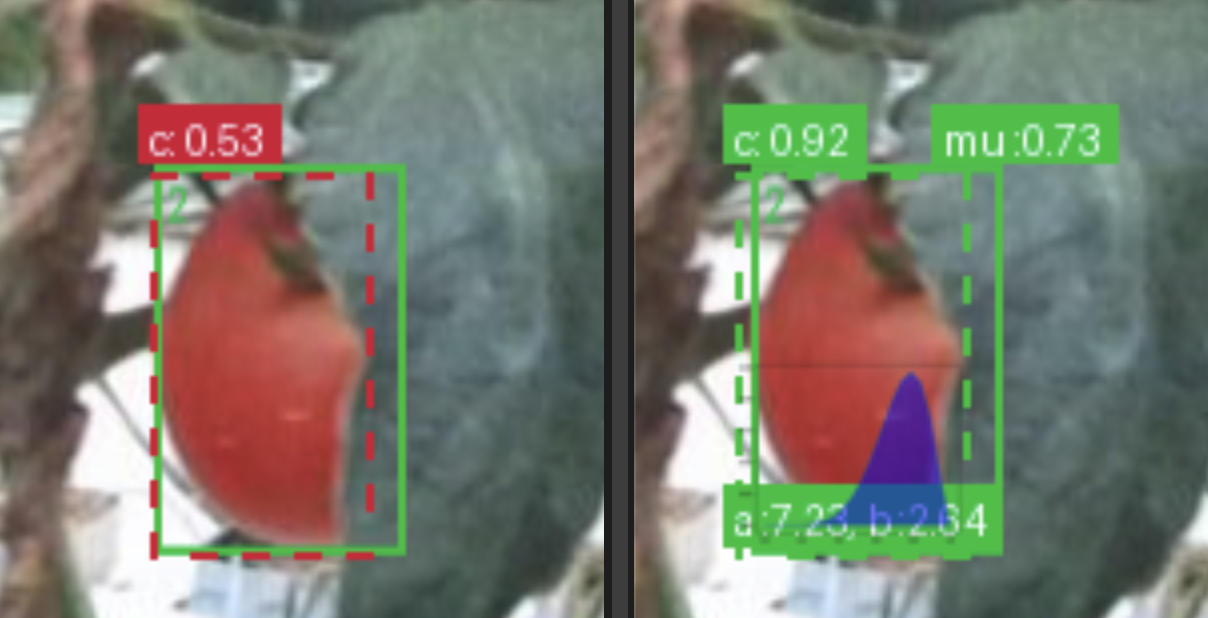}\hfill
  \includegraphics[width=0.32\linewidth]{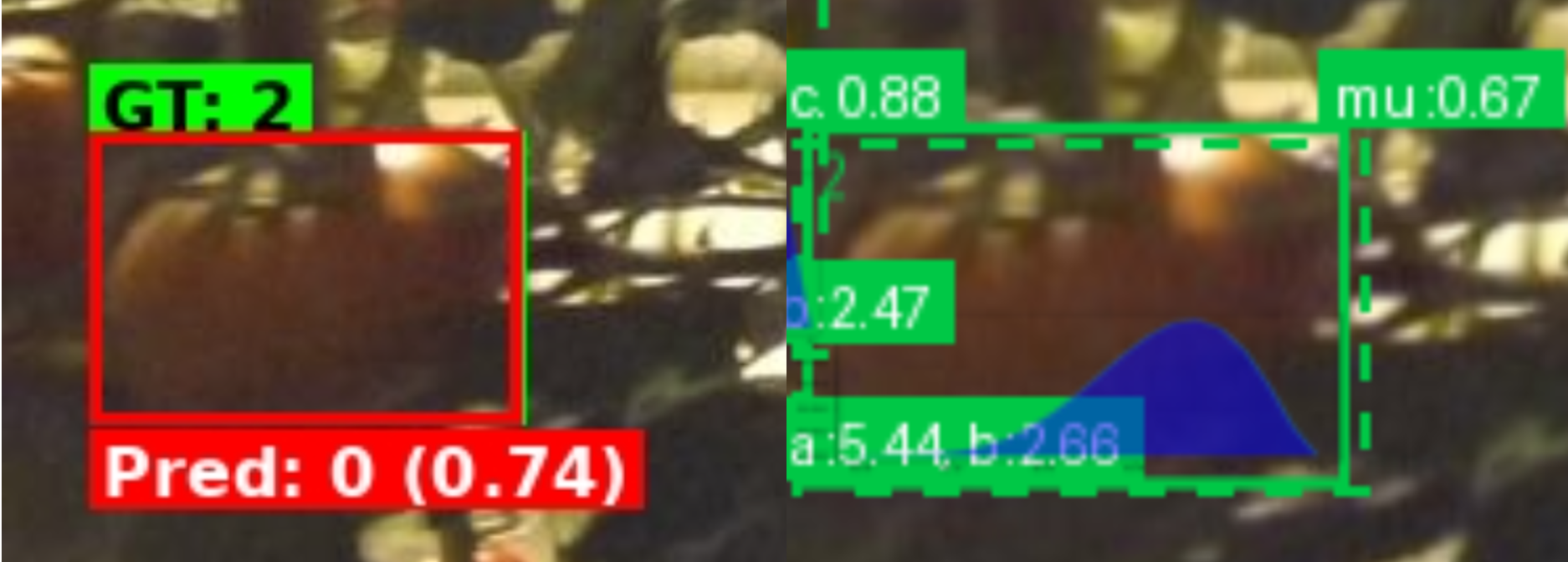}\hfill
  \includegraphics[width=0.32\linewidth]{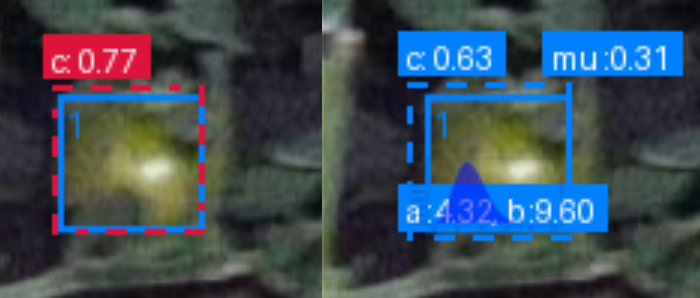}
  \vspace{-3mm} 
  \caption{Qualitative comparison on transitional samples. Baseline (left of each pair) struggles with conflicting boxes or incorrect classes. FruitProm (right) replaces discrete failures with robust, continuous Beta distributions for maturity estimation.}
  \vspace{-4mm} 
  \label{fig:qual_results_strip}
\end{figure*}

Figure~\ref{fig:qual_results_strip} provides a qualitative illustration of this difference: on a transitional sample, the deterministic baseline produces a conflicting prediction with low confidence, whereas FruitProm yields a single high-confidence detection accompanied by an interpretable maturity distribution.


\begin{table}[htbp]
\centering
\caption{Normalized Confusion Matrices for Tomato Maturity Estimation Models}
\label{tab:ml_confusion_matrices}

\resizebox{\columnwidth}{!}{%
\begin{tabular}{@{}llccc@{}}
\toprule
& & \multicolumn{3}{c}{\textbf{Predicted Stage (\%)}} \\
\cmidrule(l){3-5} 
\textbf{Model} & \textbf{True Label} & \textbf{Immature} & \textbf{Intermediate} & \textbf{Ripe} \\ \midrule

\textbf{RT-DETRv2 Baseline} 
& Immature & \textbf{99.8} & 0.1 & 0.1 \\
& Int.     & 0.7 & \textbf{98.8} & 0.4 \\
& Ripe     & 0.0 & 0.7 & \textbf{99.3} \\ \midrule

\textbf{RT-DETRv2 Noise 10} 
& Immature & \textbf{99.2} & 0.7 & 0.1 \\
& Int.     & 0.8 & \textbf{98.5} & 0.8 \\
& Ripe     & 0.0 & 1.8 & \textbf{98.2} \\ \midrule

\textbf{FruitProm Noise 0} 
& Immature & \textbf{99.7} & 0.3 & 0.1 \\
& Int.     & 1.3 & \textbf{97.3} & 1.5 \\
& Ripe     & 0.0 & 1.5 & \textbf{98.5} \\ \midrule

\textbf{FruitProm Noise 10} 
& Immature & \textbf{99.7} & 0.3 & 0.1 \\
& Int.     & 1.3 & \textbf{97.3} & 1.5 \\
& Ripe     & 0.0 & 1.5 & \textbf{98.5} \\ \midrule

\textbf{YOLO 8L Noise} 
& Immature & \textbf{99.1} & 0.9 & 0.1 \\
& Int.     & 0.9 & \textbf{98.5} & 0.6 \\
& Ripe     & 0.0 & 5.9 & \textbf{94.1} \\ \midrule

\textbf{YOLO 8L Baseline} 
& Immature & \textbf{99.9} & 0.1 & 0.1 \\
& Int.     & 0.3 & \textbf{99.1} & 0.5 \\
& Ripe     & 0.1 & 0.8 & \textbf{99.1} \\ \midrule

\textbf{YOLO 10x Noise} 
& Immature & \textbf{98.7} & 1.2 & 0.1 \\
& Int.     & 1.0 & \textbf{98.4} & 0.6 \\
& Ripe     & 0.0 & 6.7 & \textbf{93.3} \\ \midrule

\textbf{YOLO 10x Baseline} 
& Immature & \textbf{99.8} & 0.1 & 0.1 \\
& Int.     & 0.5 & \textbf{98.7} & 0.7 \\
& Ripe     & 0.0 & 0.5 & \textbf{99.5} \\ \midrule

\textbf{YOLO 11L Noise} 
& Immature & \textbf{99.7} & 0.2 & 0.1 \\
& Int.     & 0.9 & \textbf{97.6} & 1.5 \\
& Ripe     & 0.0 & 3.0 & \textbf{97.0} \\ \midrule

\textbf{YOLO 11L Baseline} 
& Immature & \textbf{99.8} & 0.2 & 0.1 \\
& Int.     & 0.3 & \textbf{99.2} & 0.5 \\
& Ripe     & 0.0 & 1.0 & \textbf{99.0} \\ \midrule

\textbf{YOLO 11x Noise} 
& Immature & \textbf{99.7} & 0.3 & 0.0 \\
& Int.     & 1.4 & \textbf{97.7} & 0.9 \\
& Ripe     & 0.0 & 7.1 & \textbf{92.9} \\ \midrule

\textbf{YOLO 11x Baseline} 
& Immature & \textbf{99.8} & 0.2 & 0.1 \\
& Int.     & 0.4 & \textbf{99.2} & 0.4 \\
& Ripe     & 0.0 & 1.0 & \textbf{99.0} \\ \midrule

\textbf{YOLO 26L Noise} 
& Immature & \textbf{99.3} & 0.6 & 0.1 \\
& Int.     & 1.0 & \textbf{98.4} & 0.6 \\
& Ripe     & 0.0 & 2.6 & \textbf{97.4} \\ \midrule

\textbf{YOLO 26L Baseline} 
& Immature & \textbf{99.8} & 0.1 & 0.1 \\
& Int.     & 0.3 & \textbf{99.3} & 0.4 \\
& Ripe     & 0.0 & 0.8 & \textbf{99.2} \\ \bottomrule

\end{tabular}%
} 
\end{table}

\section{Discussion}\label{sec:discussion}
\subsection{Annotation Reliability}\label{sec:discussion_reliability}
The contrast between near-perfect agreement at physiological extremes and $\sim$50\% agreement at the intermediate stage reveals that label ambiguity is not uniformly distributed but concentrated at transition boundaries. This variance should not be attributed to annotator error; it reflects the continuous nature of ripening compressed into discrete bins. When annotators are forced to assign mutually exclusive categories to a continuous visual spectrum, subjective bias inevitably governs the decision boundary. This was particularly evident in the consistent tendency of both annotators to classify late-transition \textit{Intermediate} samples as \textit{Ripe}, suggesting that perceived maturity cues are weighted differently near the end of the ripening cycle.\par
It is also worth noting that both annotators reside within a small geographic radius, which may expose them to similar produce from overlapping retail sources. This shared visual exposure to locally available tomato varieties could contribute to the relatively high baseline agreement at polar stages while simultaneously producing correlated subjective thresholds at the intermediate boundary. Future studies would benefit from recruiting annotators across diverse geographic regions to disentangle perceptual consensus from shared environmental exposure.\par
The asymmetric inter-annotator confusion between \textit{Ripe} and \textit{Unripe} (6.4\% vs.\ 0.2\%) warrants comment. Both baseline-vs-annotator matrices show this confusion at $\leq$1.6\%, indicating that individual annotators rarely confuse polar extremes against a common reference. The elevated rate in the direct inter-annotator comparison likely reflects a matching artifact: in images with densely clustered fruits at different maturity stages, the IoU-based Hungarian assignment may associate spatially adjacent but physiologically distinct instances, inflating apparent cross-spectrum disagreement. Future studies should consider instance-level identity tracking or higher IoU thresholds in cluttered scenes.\par
These findings have direct implications for model training. Standard cross-entropy losses treat categorical labels as certain and mutually exclusive, forcing networks to learn decision boundaries that are, in the intermediate regime, largely artifacts of human subjectivity rather than reflections of objective physiological states. The systematic, adjacent-stage nature of the observed disagreement motivates our proposed FruitProm architecture: by predicting continuous $\alpha$--$\beta$ distribution parameters, the model can explicitly represent the latent uncertainty that discrete labels fail to capture.

\subsection{Detection Performance and Noise Robustness}
\label{sec:discussion_detection}
On the clean dataset, the CNN-based YOLO family achieves the highest raw mAP, reflecting the well-documented strength of these architectures on small-to-medium-scale detection benchmarks. The FruitProm model, built on the transformer-based RT-DETRv2 backbone, trails the best YOLO variant by approximately 3 points in mAP\textsubscript{50--95} (0.845 vs.\ 0.874). This gap is expected: the probabilistic head introduces a distributional parameterization that trades marginal clean-data accuracy for a fundamentally different supervision signal. The critical finding emerges under label noise. All CNN-based models suffer a 3.5--4.5\% relative mAP drop at 10\% symmetric noise, with the degradation concentrated in the \textit{Ripe}$\rightarrow$\textit{Intermediate} confusion cell---precisely the transition boundary where our annotation study identified the greatest human disagreement. This confirms that deterministic cross-entropy supervision amplifies the very label ambiguity that exists in real-world agricultural datasets. FruitProm, by contrast, exhibits near-complete noise invariance: a 0.59\% relative drop, and---notably---\textit{identical} per-class confusion matrices under clean and noisy conditions. This invariance arises because the CDF-based focal loss treats each label as an interval observation on a continuous maturity axis rather than a hard categorical target, preventing noisy boundary labels from distorting the learned distribution. An architectural distinction also contributes to this robustness gap. The CNN-based YOLO models rely on local convolutional receptive fields that are sensitive to per-instance label assignments, whereas the transformer-based RT-DETRv2 backbone benefits from global self-attention that can contextualize maturity cues across the full image. Even the deterministic RT-DETRv2 baseline shows a smaller drop ($-$3.07\%) than any YOLO variant, suggesting that the attention mechanism provides a degree of implicit noise resilience. Our probabilistic head builds on this architectural advantage, combining the transformer's contextual reasoning with explicit distributional uncertainty modeling. Although FruitProm can in principle be applied to any detection backbone, we elected not to pursue CNN-based instantiations in this study. Given that the YOLO models already exhibited substantial sensitivity to label noise under deterministic supervision, the more productive path was to pair the probabilistic formulation with the architecture that provides the strongest foundation for noise-robust inference.
\section{Conclusion}\label{sec:conclusion}
We presented FruitProm, a probabilistic reformulation of fruit maturity estimation that replaces deterministic categorical classification with a continuous $\alpha$--$\beta$ distribution head supervised via a CDF-based focal loss. An inter-annotator reliability study confirmed that human labeling uncertainty is systematically concentrated at maturity transition boundaries, empirically validating the need to move beyond discrete class assumptions. On the Tomato Maturity benchmark, FruitProm preserves competitive detection performance on clean data while exhibiting near-complete invariance to 10\% symmetric label noise---reducing the relative mAP drop from 3--4.5\% (deterministic baselines) to 0.59\%. These results demonstrate that explicitly modeling maturity as a latent continuous process yields representations that are both more faithful to the underlying biology and substantially more robust to the annotation ambiguity inherent in real-world agricultural datasets.

{
    \small
    \bibliographystyle{ieeenat_fullname}
    \bibliography{main}
}
\end{document}